\documentclass{article}
\usepackage{ijcai25}
\usepackage{times}
\usepackage{soul}
\usepackage{makecell}
\usepackage{url}
\usepackage[hidelinks]{hyperref}
\usepackage[small]{caption}
\usepackage{graphicx}
\usepackage{amsmath}
\usepackage{amsthm}
\usepackage{booktabs}
\usepackage{algorithm}
\usepackage{algorithmic}
\usepackage[switch]{lineno}
\usepackage{hyperref}
\usepackage{svg} 
\usepackage{tikz}
\usepackage{float}
\usepackage{amssymb}
\usetikzlibrary{trees, shapes, positioning}
\usepackage{fontawesome5}
\usepackage{xcolor}
\newcommand{\linkicon}{{\color{blue}\faLink}}

\title{Autoformalization in the Era of Large Language Models: A Survey}

\author{
Ke Weng$^1$\thanks{Equal Contribution}
\and
Lun Du$^2$\footnotemark[1]\thanks{Corresponding Author}\and
Sirui Li$^1$\and
Wangyue Lu$^1$\and
Haozhe Sun$^1$ \and
Hengyu Liu$^3$\And\\
Tiancheng Zhang$^1$\footnotemark[2]\\
\affiliations
$^1$Northeastern University\\
$^2$Ant Research Institute, Ant Group\\
$^3$Department of Computer Science, Aalborg University\\
\emails
\{2401925, 20226504, 2410827, 20223246\}@stu.neu.edu.cn,\\
dulun.dl@antgroup.com, heli@cs.aau.dk, tczhang@mail.neu.edu.cn 
}

\begin{document}

\maketitle

\begin{abstract}
Autoformalization—the process of transforming informal mathematical propositions into verifiable formal representations—is a foundational task in automated theorem proving, offering a new perspective on the use of mathematics in both theoretical and applied domains. Driven by the rapid progress in artificial intelligence, particularly large language models (LLMs), this field has witnessed substantial growth, bringing both new opportunities and unique challenges. In this survey, we provide a comprehensive overview of recent advances in autoformalization from both mathematical and LLM-centric perspectives. We examine how autoformalization is applied across various mathematical domains and levels of difficulty, and analyze the end-to-end workflow—from data preprocessing to model design and evaluation. We further explore the emerging role of autoformalization in enhancing the verifiability of LLM-generated outputs, highlighting its potential to improve both the trustworthiness and reasoning capabilities of LLMs. Finally, we summarize key open-source models and datasets supporting current research, and discuss open challenges and promising future directions for the field.
\end{abstract}

\begin{table*}[h]
\centering
\caption{Open source model or method with link and brief description. The superscripts in the ``Dataset / Benchmark'' column indicate the following: \textsuperscript{1} means the dataset was used for evaluation in the paper, \textsuperscript{2} means it was used for training, and \textsuperscript{*} denotes that the dataset was first proposed in that paper. }
\label{tab:my-table}
\resizebox{\textwidth}{!}{%
\begin{tabular}{lll}
\toprule
Method / Model \& Link   & Dataset / Benchmark & Method / Model Tasks  \\ \midrule

DeepSeek-Prover-V1\textsuperscript{\href{https://huggingface.co/deepseek-ai/DeepSeek-Prover-V1}{\linkicon}}            
    & MiniF2F\textsuperscript{1,2}& Formal Proof Gen, Data Construction  \\

LeanEuclid\textsuperscript{\href{https://github.com/loganrjmurphy/LeanEuclid}{\linkicon}}
    & LeanEuclid\textsuperscript{2,*}, Elements\textsuperscript{1}, UniGeo\textsuperscript{1} & Formal Proof Gen, Autoformalization \\

Isa-AutoFormal\textsuperscript{\href{https://github.com/Miracle-Messi/Isa-AutoFormal}{\linkicon}}
    & Math\textsuperscript{1}, MiniF2F\textsuperscript{1} & Autoformalization, Informalization \\

Retrieval Augmented autoformalization\textsuperscript{\href{https://github.com/lanzhang128/retrieval_augmented_autoformalization}{\linkicon}}
    & IsarMathLib\textsuperscript{2} & RAG, Autoformalization \\

DeepSeekMath\textsuperscript{\href{https://github.com/deepseek-ai/DeepSeek-Math}{\linkicon}}
    & \makecell[l]{GSM8K\textsuperscript{1}, MATH\textsuperscript{1}, SAT\textsuperscript{1}, OCW Courses\textsuperscript{1}, MGSM-zh\textsuperscript{1}, \\ CMATH\textsuperscript{1}, Gaokao-MathCloze\textsuperscript{1}, Gaokao-MathQA\textsuperscript{1}, MiniF2F\textsuperscript{1}} & Formal Proof Gen \\

DRAFT, SKETCH, AND PROVE\textsuperscript{\href{https://github.com/albertqjiang/draft_sketch_prove}{\linkicon}}
    & MiniF2F\textsuperscript{1} & Formal Proof Gen \\

FORMALALIGN\textsuperscript{\href{https://github.com/rookie-joe/FormalAlign}{\linkicon}}
    & MiniF2F\textsuperscript{1}, FormL4\textsuperscript{1,2}, MMA\textsuperscript{2} & Autoformalization and Evaluation \\

GFLEAN\textsuperscript{\href{https://github.com/pkshashank/GFLeanTransfer?tab=readme-ov-file}{\linkicon}}
    & textbook Mathematical Proofs\textsuperscript{1} & AST simplification and transformation \\

Lean Workbook\textsuperscript{\href{https://github.com/InternLM/InternLM-Math?tab=readme-ov-file}{\linkicon}}
    & MiniF2F\textsuperscript{2}, ProofNet\textsuperscript{2} & Data Construction \\

LEAN-GitHub\textsuperscript{\href{https://github.com/InternLM/InternLM-Math?tab=readme-ov-file}{\linkicon}}
    & MiniF2F\textsuperscript{1}, ProofNet\textsuperscript{1}, Putnam\textsuperscript{1} & Data Construction \\

LeanDojo\textsuperscript{\href{https://github.com/lean-dojo}{\linkicon}}
    & LeanDojo\textsuperscript{1,2,*} & RAG, Formal Proof Gen \\

LEGO-PROVER\textsuperscript{\href{https://github.com/wiio12/LEGO-Prover}{\linkicon}}
    & MiniF2F\textsuperscript{1,2,*} & RAG, Formal Proof Gen \\

LLEMMA\textsuperscript{\href{https://github.com/EleutherAI/math-lm}{\linkicon}}
    & Math\textsuperscript{1}, GSM8K\textsuperscript{1}, MiniF2F\textsuperscript{1}, ProofFile\textsuperscript{2,*} & Formal Proof Gen \\

LYRA\textsuperscript{\href{https://github.com/chuanyang-Zheng/Lyra-theorem-prover}{\linkicon}}
    & MiniF2F\textsuperscript{1,2} & Formal Proof Gen, Autoformalization \\

MMA\textsuperscript{\href{https://github.com/albertqjiang/mma}{\linkicon}}
    & MiniF2F\textsuperscript{1}, ProofNet\textsuperscript{1}, MMA\textsuperscript{2,*} & Informalization, Data Construction \\

Pantograph\textsuperscript{\href{https://github.com/stanford-centaur/PyPantograph?tab=readme-ov-file}{\linkicon}}
    & MiniF2F\textsuperscript{1} & Formal Proof Gen, Data Construction \\

PDA\textsuperscript{\href{https://github.com/rookie-joe/PDA}{\linkicon}}
    & FormL4\textsuperscript{1,2,*} & Data Construction, Informalization \\

ProofNet\textsuperscript{\href{https://github.com/zhangir-azerbayev/ProofNet?tab=readme-ov-file}{\linkicon}}
    & ProofNet\textsuperscript{1,2,*} & RAG, Informalization \\

SubgoalXL\textsuperscript{\href{https://github.com/zhaoxlpku/SubgoalXL}{\linkicon}}
    & MiniF2F\textsuperscript{1,2} & Formal Proof Gen \\

Herald\textsuperscript{\href{https://huggingface.co/FrenzyMath/Herald_translator}{\linkicon}}
    & MiniF2F\textsuperscript{1}, Extract Theorem\textsuperscript{1}, College CoT\textsuperscript{1} & RAG, Data Construction, Autoformalization \\

DeepSeek-Prover-V2\textsuperscript{\href{https://github.com/deepseek-ai/DeepSeek-Prover-V2}{\linkicon}}
    & MiniF2F\textsuperscript{1}, ProofNet\textsuperscript{1}, Putnam\textsuperscript{1}, CombiBench\textsuperscript{1}, ProverBench\textsuperscript{1,*} & Formal Proof Gen, RAG, Data Construction \\

Kimina-Prover\textsuperscript{\href{https://github.com/MoonshotAI/Kimina-Prover-Preview}{\linkicon}}
    & MiniF2F\textsuperscript{1}  & Formal Proof Gen, Autoformalization, Data Construction \\

Goedel-Prover\textsuperscript{\href{https://github.com/Goedel-LM/Goedel-Prover}{\linkicon}}
    & MiniF2F\textsuperscript{1}, Putnam\textsuperscript{1}, ProofNet\textsuperscript{1}, Lean Workbook\textsuperscript{1,2} & Formal Proof Gen  \\

STP\textsuperscript{\href{https://github.com/kfdong/STP}{\linkicon}}
    & MiniF2F\textsuperscript{1}, Putnam\textsuperscript{1}, ProofNet\textsuperscript{1} & Formal Proof Gen  \\

Leanabell-Prover\textsuperscript{\href{https://github.com/Leanabell-LM/Leanabell-Prover}{\linkicon}}
    & MiniF2F\textsuperscript{1} & Formal Proof Gen, Data Construction  \\

BFS-Prover\textsuperscript{\href{https://huggingface.co/ByteDance-Seed/BFS-Prover}{\linkicon}}
    & MiniF2F\textsuperscript{1} & Formal Proof Gen, RAG, Data Construction  \\

InternLM2.5-StepProver\textsuperscript{\href{https://github.com/InternLM/InternLM-Math}{\linkicon}}
    & MiniF2F\textsuperscript{1}, Putnam\textsuperscript{1}, ProofNet\textsuperscript{1}, Lean-Workbook\textsuperscript{1,*}  & Formal Proof Gen, Data Construction \\

LoT-Solver\textsuperscript{\href{https://huggingface.co/RickyDeSkywalker/LoT-Solver}{\linkicon}}
    & MiniF2F\textsuperscript{1}  & Formal Proof Gen \\

PDE-Controller\textsuperscript{\href{https://github.com/delta-lab-ai/pde-controller}{\linkicon}}
    & pde-controller\textsuperscript{1,2,*}& Formal Proof Gen, Data Construction, Autoformalization \\

DeepTheorem\textsuperscript{\href{https://huggingface.co/Jiahao004/DeepTheorem-qwen-7b-rl}{\linkicon}}
    & DeepTheorem\textsuperscript{1,2,*}& Formal Proof Gen, Data Construction \\

REAL-Prover\textsuperscript{\href{https://github.com/frenzymath/REAL-Prover}{\linkicon}}
    & FATE-M\textsuperscript{2,*}, ProofNet\textsuperscript{2}, MiniF2F\textsuperscript{2}, Numinamath\textsuperscript{1}, Lean-Workbook\textsuperscript{1}& Formal Proof Gen, Data Construction, Autoformalization \\

\bottomrule
\end{tabular}%
}
\end{table*}

\section{Introduction}
Throughout history, mathematicians have relied on various tools-mechanical, or electronic—to assist their work, from early devices like the abacus to modern computing systems  \cite{tao2024machine}. In recent decades, formal proof assistants such as Lean \cite{moura2021lean}, Coq \cite{huet1997coq}, and Isabelle \cite{paulson1994isabelle} have emerged as powerful environments for constructing mathematically rigorous proofs with machine-checkable correctness. Unlike traditional programming languages, formal languages are not designed to perform computations but to encode and verify the logical validity of mathematical arguments within a formal system. Each step of a proof typically expands into multiple lines of formal code, and only when the entire script complies with the rules of logic will the proof be accepted as correct.

Despite their capabilities, using proof assistants remains highly labor-intensive, even for experts, since formalizing informal mathematical reasoning requires precision at every logical step. This challenge has given rise to \textit{autoformalization}, a research direction focused on automating the translation from informal mathematics to formal language. As a key component of the broader field of mathematical artificial intelligence, autoformalization seeks to bridge the gap between human intuition and machine-verifiable reasoning.

Recent breakthroughs in large language models (LLMs) have accelerated progress in this direction. Notably, AlphaProof \cite{alphaproof2024ai} demonstrated silver-medal-level performance on International Mathematical Olympiad (IMO) problems, showcasing the potential of LLMs in tackling high-level mathematical reasoning tasks. More recently, GPT-o3 reportedly solved 25.2\% of problems in the Frontier Mathematics benchmark, significantly outperforming earlier models that achieved only around 2\%. These milestones highlight the growing ability of LLMs to generate complex mathematical solutions, while also underscoring the need for formal verification to ensure correctness.


Beyond mathematics, however, the core principles of autoformalization—translating informal or semi-formal language into precise, verifiable representations—hold broader significance. As LLMs are increasingly applied to tasks like algorithm design, code generation, and system specification \cite{shi2023sotana,di2024codefuse,zheng2025knowledge}, ensuring that their outputs are not only syntactically correct but also semantically sound becomes critical. Autoformalization offers a promising mechanism for bridging this gap: by converting natural language specifications, pseudo-code, or even informal correctness claims into formal logic or verifiable code contracts, it enables rigorous checking of algorithmic behavior. In domains such as software engineering, security auditing, and scientific computing, this capability could substantially enhance the trustworthiness and reliability of LLM-generated artifacts. Viewed from this perspective, autoformalization is not merely a tool for mathematical proof—it is a foundational enabler for building LLM systems that can reason, code, and verify in an integrated and principled way.

This survey provides a comprehensive and structured review of the recent advances in automatic formalization, focusing on the current applications in mathematical theorem proving and the possibility of combining automatic formalization with formal verification of large models to generate fully trustworthy outputs in the future. §2 shows a review of the historical and technological foundations of autoformalization. And we summarize notable open-source models and datasets that support current research in §3; In §4, we dissect the autoformalization workflow, covering key components such as data preprocessing, modeling strategies, postprocessing techniques, and evaluation protocols, and also dissect state of the art model approaches to mathematical theorem proving. In §5, we conduct a deep analysis of the involved mathematical problems from the mathematical problem domain, the mathematical problem difficulty level and the mathematical problem abstraction level. In §6, we explore the connection between autoformalization and the formal verification of large model output from multiple fields. In §7, we discuss major open challenges and promising future research directions. Finally, §8 compares our survey with existing related surveys, clarifying its unique perspective and contributions.

\begin{table*}[h]
\centering
\caption{Datasets/Benchmarks with code repos link and description}
\label{tab:your-table}
\resizebox{\textwidth}{!}{%
\begin{tabular}{ccc}
\toprule
Dataset/Benchmark \& Link &Size   & Description \\ \midrule
\href{https://github.com/openai/miniF2F}{MiniF2F} &488   & Statements from olympiads as well as high-school and undergraduate maths classes.\\
\href{https://github.com/loganrjmurphy/LeanEuclid}{LeanEuclid} & 173 &Euclidean geometry problems manually formalized in Lean.\\
\href{https://github.com/hendrycks/math}{Math}& 12,500 & 12,500 challenging competition mathematics problems.\\
\href{https://github.com/rookie-joe/PDA}{FORML4}& 17,137 & Informalize theorems extracted from Mathlib 4. \\
\href{https://github.com/rahul3613/ProofNet-lean4}{ProofNet}&374 & The problems are drawn from popular undergraduate pure mathematics textbooks. \\
\href{https://github.com/trishullab/PutnamBench}{PutnamBench}&657 & Competition mathematics problems sourced from the William Lowell Putnam Mathematical Competition. \\
\href{https://zenodo.org/records/12740403}{LeanDojo}& 122,517   & Theorems,proofs,tactics,premises from mathlib4 \\
\href{https://huggingface.co/datasets/openai/gsm8k}{GSM8K}& 8,792   & High quality linguistically diverse grade school math word problems.\\
\href{https://huggingface.co/datasets/l3lab/miniCTX}{miniCTX}&762  & Theorems sourced from real Lean projects and textbooks.\\
\href{https://github.com/jlab-nlp/arxiv2formal}{arxiv2Formal}&50  & Theorems from arXiv papers formalized in Lean.\\    
\href{https://huggingface.co/datasets/deepseek-ai/DeepSeek-Prover-V1}{DeepSeek-Prover-V1}& 27,503  & Lean4 proof data derived from high-school and undergraduate-level mathematical competition problems.\\
\href{https://huggingface.co/datasets/internlm/Lean-Workbook}{Lean-Workbook}& 13,517  & Collection of mathematical problems and their proof information.\\
\href{https://huggingface.co/datasets/internlm/Lean-Github}{Lean-GitHub}& 20,446  &  Theorems with formal proofs and tactics from 2,133 files. \\                 \href{https://github.com/dwrensha/compfiles}{Compfiles}  &205  & A collection of olympiad-style math problems and their solutions, formalized in Lean 4.\\
\href{https://huggingface.co/datasets/FrenzyMath/Herald_statements}{Herald}&624,436  & The dataset generated from Herald pipeline on Mathlib4, containing 580k valid statements and 44k NL-FL proofs.\\
\href{https://huggingface.co/datasets/deepseek-ai/DeepSeek-ProverBench}{ProverBench}&325  & A benchmark dataset containing 325 formalized problems to advance neural theorem proving research, including 15 from the prestigious AIME competitions (years24-25).\\
\href{https://github.com/MoonshotAI/CombiBench}{CombiBench}&100  & A manually produced benchmark, including 100 combinatorial mathematics problems of varying difficulty and knowledge levels.\\
\href{https://huggingface.co/datasets/kfdong/STP_Lean_0320}{STP\_Lean\_0320}&3262558 & A dataset including mathlib4 examples, LeanWorkbook proofs, and self-play conjecture proofs.\\
\href{https://huggingface.co/datasets/stoney0062/Leanabell-Prover-Formal-Statement}{Leanabell-Prover-Formal-Statement }&1,133,283  & A dataset including released data by existing models and synthetic data from informal math problems.\\
\href{https://huggingface.co/datasets/zzzzzhy/Ineq-Comp}{IneqComp}&75   & A benchmark built from elementary inequalities through systematic transformations, including variable duplication, algebraic rewriting, and multi-step composition.\\
\href{https://huggingface.co/datasets/HuajianXin/APE-Bench_I}{APE-Bench I}&10928   & The first large-scale, file-level benchmark for proof engineering, built from over 10,000 real-world commits in Mathlib4.\\
\href{https://huggingface.co/SphereLab}{Formal Math}&5560   & A benchmark of 5,560 formally verified mathematical statements spanning diverse subdomains. This dataset is 22.8× larger than the widely used MiniF2F benchmark.\\
\href{https://huggingface.co/datasets/Jiahao004/DeepTheorem}{DeepTheorem}&121000   & It comprises 121K IMO-level informal theorems and proofs spanning diverse mathematical domains.\\
\href{https://huggingface.co/collections/AI-MO/numinamath-6697df380293bcfdbc1d978c}{Numinamath}&860000   & A dataset of math competition problem-solution pairs, with each solution templated with Chain of Thought (CoT) reasoning.\\

\bottomrule
\end{tabular}%
}
\end{table*}

\begin{table}[h]
    \centering
    \caption{Availability of Informal/Formal Statements and Solutions Across Datasets}
    \label{tab:your-table2}
    \tiny
    \setlength{\tabcolsep}{2pt} 
    {%
    \begin{tabular}{lcccc}
    \toprule
        Dataset \& Link & Informal Statement & Formal Statement& Informal Solution & Formal Solution\\ \midrule
        MiniF2F~\textsuperscript{\href{https://github.com/openai/miniF2F}{\linkicon}} & &\checkmark & &\checkmark \\

         LeanEuclid\textsuperscript{\href{https://github.com/loganrjmurphy/LeanEuclid}{\linkicon}}
                      & \checkmark & \checkmark &  & \checkmark\\
         Math\textsuperscript{\href{https://github.com/hendrycks/math}{\linkicon}}
                      & \checkmark &  & \checkmark & \\
    FORML4\textsuperscript{\href{https://github.com/rookie-joe/PDA}{\linkicon}}
                      & \checkmark & \checkmark & \checkmark & \checkmark\\
    ProofNet\textsuperscript{\href{https://github.com/rahul3613/ProofNet-lean4}{\linkicon}}
                      & \checkmark & \checkmark & \checkmark & \\
    PutnamBench\textsuperscript{\href{https://github.com/trishullab/PutnamBench}{\linkicon}}
                      & \checkmark &  & \checkmark & \\

    LeanDojo\textsuperscript{\href{https://zenodo.org/records/12740403}{\linkicon}}
                      &  &  &  & \checkmark\\
    GSM8K\textsuperscript{\href{https://huggingface.co/datasets/openai/gsm8k}{\linkicon}}
                      &  &  &  & \\
    miniCTX\textsuperscript{\href{https://huggingface.co/datasets/l3lab/miniCTX}{\linkicon}}
                      &  & \checkmark &  & \checkmark\\
    arxiv2Formal\textsuperscript{\href{https://github.com/jlab-nlp/arxiv2formal}{\linkicon}}
                      & \checkmark & \checkmark &  & \\
    DeepSeek-Prover-V1\textsuperscript{\href{https://huggingface.co/datasets/deepseek-ai/DeepSeek-Prover-V1}{\linkicon}}
                      &  & \checkmark &  & \checkmark\\
    Lean-Workbook\textsuperscript{\href{https://huggingface.co/datasets/internlm/Lean-Workbook}{\linkicon}}
                      & \checkmark & \checkmark &  & \checkmark\\

    Lean-GitHub\textsuperscript{\href{https://huggingface.co/datasets/internlm/Lean-Github}{\linkicon}}
                      &  &  &  & \checkmark\\
    Compfiles\textsuperscript{\href{https://github.com/dwrensha/compfiles}{\linkicon}}
                      &  & \checkmark &  & \checkmark\\
    Herald\textsuperscript{\href{https://huggingface.co/datasets/FrenzyMath/Herald_statements}{\linkicon}}
                      & \checkmark & \checkmark & \checkmark & \checkmark\\
    ProverBench\textsuperscript{\href{https://huggingface.co/datasets/deepseek-ai/DeepSeek-ProverBench}{\linkicon}}
                      &  & \checkmark &  & \\
    CombiBench\textsuperscript{\href{https://github.com/MoonshotAI/CombiBench}{\linkicon}}
                      & \checkmark & \checkmark &  & \\

    STP\_Lean\textsuperscript{\href{https://huggingface.co/datasets/kfdong/STP_Lean_0320}{\linkicon}}
                      &  & \checkmark &  & \checkmark\\
    Leanabell\textsuperscript{\href{https://huggingface.co/datasets/stoney0062/Leanabell-Prover-Formal-Statement}{\linkicon}}
                      & \checkmark & \checkmark &  & \\
    IneqComp\textsuperscript{\href{https://huggingface.co/datasets/zzzzzhy/Ineq-Comp}{\linkicon}}
                      &  & \checkmark &  & \checkmark\\
    APE-Bench I\textsuperscript{\href{https://huggingface.co/datasets/HuajianXin/APE-Bench_I}{\linkicon}}
                      & \checkmark & \checkmark &  & \\
    Formal Math\textsuperscript{\href{https://huggingface.co/SphereLab}{\linkicon}}
                      & \checkmark & \checkmark & \checkmark & \\
    DeepTheorem\textsuperscript{\href{https://huggingface.co/datasets/Jiahao004/DeepTheorem}{\linkicon}}
                      & \checkmark & \checkmark & \checkmark & \\
    Numinamath\textsuperscript{\href{https://huggingface.co/collections/AI-MO/numinamath-6697df380293bcfdbc1d978c}{\linkicon}}
                      & \checkmark &  & \checkmark & \\
    \bottomrule
    \end{tabular}
    }
\end{table}

\section{Background}
Autoformalization plays a dual role in contemporary AI and mathematical reasoning. On one hand, it facilitates the formalization of mathematical knowledge, accelerating the development of machine-verified proofs and theorem-proving systems. On the other, it offers a framework for grounding the outputs of large language models (LLMs) in formal logic, enabling verifiability and fostering deeper integration between symbolic and neural methods \cite{wu2022autoformalizationlargelanguagemodels}. This synergy creates a positive feedback loop: machine learning advances assist in formalization, while formal systems, in turn, provide rigorous tools for verifying and improving model outputs.

In this section, we introduce the core concepts of autoformalization from two complementary perspectives: its foundational role in mathematical formalization, and its emerging importance in the formal verification of LLM-generated content.

\subsection{Autoformalization in Mathematics}
In recent years, mathematicians have successfully completed the formalization of several landmark theorems, including the Four Color Theorem \cite{Gonthier2008FormalPF} and Kepler's Conjecture \cite{hales2015formalproofkeplerconjecture}. However, the formalization process remains highly labor-intensive and time-consuming. For instance, formalizing the Four Color Theorem took nearly five years, while Kepler’s Conjecture required more than a decade of effort.

Autoformalization seeks to alleviate this burden by automating the translation of informal proofs into representations verifiable by proof assistants such as Coq \cite{Coq}, Lean \cite{10.1007/978-3-319-21401-6_26}, and Isabelle \cite{mohamed2008theorem}. By reducing manual formalization effort, autoformalization enables mathematicians to devote more attention to conceptual reasoning and mathematical insight, while still benefiting from the rigor and reliability provided by formal verification.

\begin{figure}[h]
    \centering
    \includegraphics[width=1\linewidth]{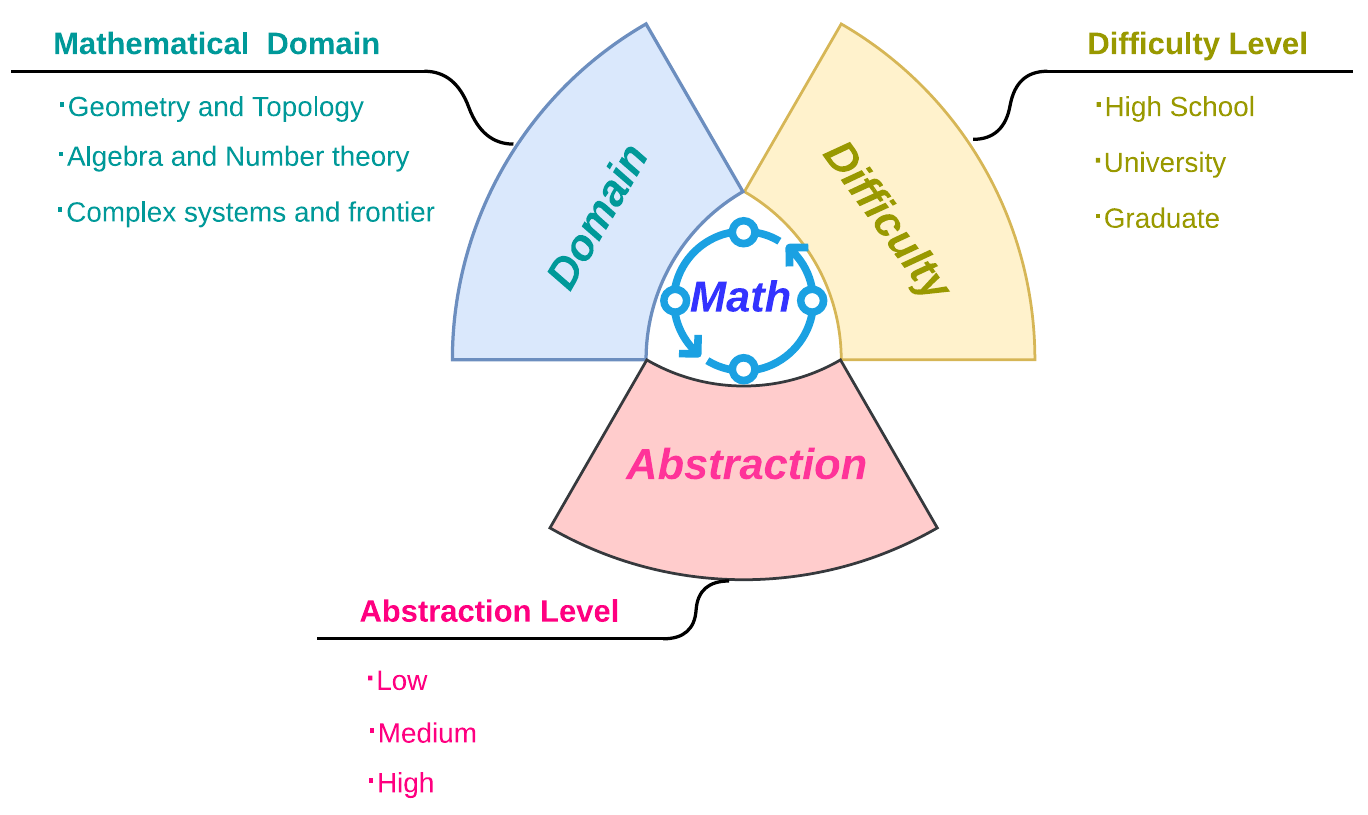}
    \caption{Multi-dimensional partitioning Autoformalization in Mathematics}
    \label{fig:enter1-label}
\end{figure}

To better understand and systematically analyze autoformalization in the mathematical domain, we propose a three-dimensional framework, as illustrated in Figure~\ref{fig:enter1-label}. This framework captures the key axes along which existing research on autoformalization can be categorized and compared:

\begin{itemize}
    \item \textbf{Mathematical Problem Domain:} Autoformalization tasks vary significantly depending on the subfields they target. We highlight representative areas such as geometry and topology, algebra and number theory, and complex systems or frontier problems that often involve interdisciplinary reasoning.
    
    \item \textbf{Difficulty Level:} We further classify problems based on their intrinsic complexity and required level of abstraction—from high school mathematics to undergraduate and graduate-level reasoning. This axis reveals how current systems cope with increasing proof complexity and conceptual depth.
    
    \item \textbf{Workflow:} Finally, we analyze the typical autoformalization pipeline, including data preprocessing, modeling approaches, post-processing methods, and evaluation strategies. This dimension offers insights into how different systems are constructed, trained, and evaluated.
\end{itemize}

We adopt this framework to guide the subsequent sections of the paper, providing a structured discussion of current research efforts along each of these three dimensions.

\subsection{Autoformalization in LLM Output Formal Verification}

The advent of increasingly powerful large language models (LLMs)—such as LLaMA \cite{touvron2023llamaopenefficientfoundation}, Claude \cite{claude}, Gemini \cite{geminiteam2024geminifamilyhighlycapable}, and GPT-4 \cite{openai2024gpt4technicalreport}—has enabled remarkable progress across a wide range of tasks, thanks to their broad domain knowledge and strong few-shot reasoning capabilities. However, despite their fluency and apparent competence, LLMs are prone to generating outputs that are factually incorrect or logically unsound—a phenomenon widely referred to as \textbf{hallucination} \cite{sadasivan2023can}. This issue is further compounded by the models' ability to produce persuasive and human-like responses, making it challenging to distinguish valid reasoning from subtle but critical errors.

To address this challenge, recent research has explored the use of autoformalization as a means of translating natural language outputs from LLMs into precise, verifiable representations using formal systems such as first-order logic \cite{ryu2024divide} or arithmetic system such as Peano arithmetic\cite{Peano}. \cite{balcan2025learningverifierschainofthoughtreasoning} is also a recent research which focus on the chain of thought by designing a formal definition verifier and training the model to learn the correctness of the reasoning sequence, aiming to ensure that each reasoning step of the chain of thought is valid. In this context, autoformalization acts as a bridge between the inherent ambiguity of natural language and the strict requirements of formal verification. By enabling symbolic reasoning over LLM outputs, autoformalization enhances correctness, improves reliability, and promotes the safe and trustworthy deployment of LLMs in tasks requiring rigorous reasoning and high assurance.

\begin{figure}[h]
    \centering
    \includegraphics[width=1\linewidth]{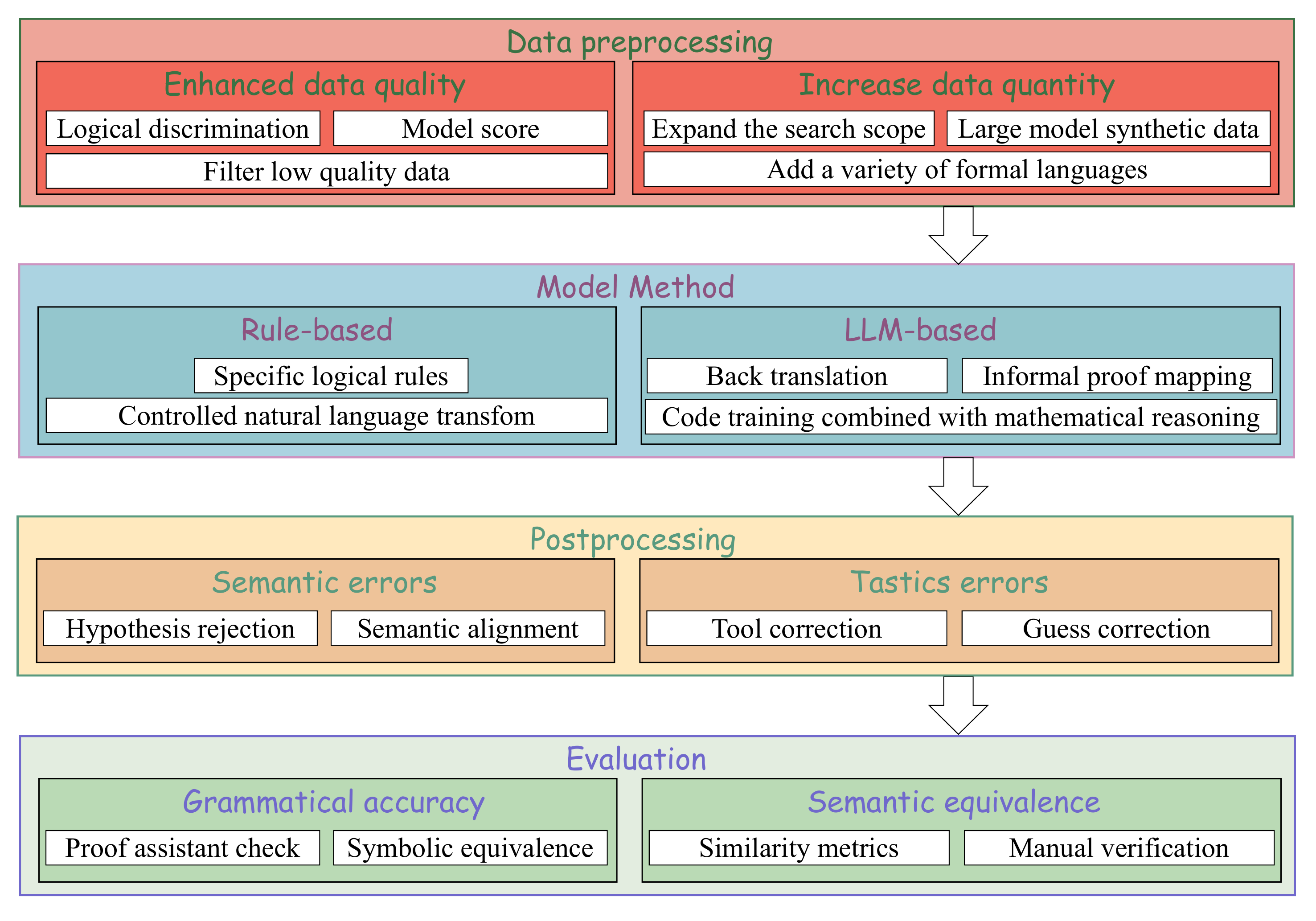}
    \caption{The framework of workflow for Autoformalization}
    \label{fig:enter2-label}
\end{figure}

\section{Foundations of Progress: Techniques and Corpora}

In order to systematically investigate the field of autoformalization, we have collected the open-source code, model parameters, and datasets from existing works, as shown in Table\ref{tab:my-table} and Table\ref{tab:your-table}. Table \ref{tab:my-table} describes the existing works with links to open source code for use. We also summarize the benchmarks used by each method/model and the role of benchmarks in the corresponding method/model. Top corner mark$^{1}$ represents the evaluation of performance, top corner mark $^{2}$ represents the training model as a dataset, and top corner mark $^{*}$ represents the benchmark proposed by the paper. We also summarized the tasks for each method/model as follows:
\begin{itemize}
    \item \textbf{Formal Proof Gen}: Using the reasoning ability of llm, the given natural language mathematical problem statement is automatically generated into formal proofs that can be verified by formal systems, and the corresponding grammar rules and solution logic are satisfied.
    \item \textbf{Data Construction}: Through large-scale synthesis and generation, enhancing or filtering data, the bottleneck of data scarcity is broken and the performance of the model is improved. It aims to systematically extract, annotate, and structure the training data required for formal model learning from existing theorem libraries, proof scripts, or natural language mathematics. The goal is to provide high-quality, large-scale parallel prediction for downstream tasks, supporting supervised learning and reinforcement learning training.
    \item \textbf{Evaluation}: Compilation of formal proof assistans does not mean that autoformalization is completely correct. In order to better verify the correctness of autoformalization, new methods or performance metrics are proposed.
    \item \textbf{Autoformalization}: It involves the automatic conversion of mathematical statements expressed in natural language into formal language statements. This task requires the model to complete a representation that can be verified by a formal system under multiple constraints such as semantic consistency, contextual completeness, and symbolic accuracy, forming a key bridge between natural language and formal systems.
    \item \textbf{Informalization}: The information in the formal statement is always enough to infer the informal statement, but the reverse is not always true. From the analysis experience, informalization is easier than formalization. Informalization refers to the process of enhancing autoformalization capabilities by studying the transformation from formal languages to informal languages.
    \item \textbf{RAG}: Given the current state of the proof, enhance the ability to retrieve potential premises and establish reflection between them. By retrieving semantically related previous proofs or structures from the theorem library, RAG assists the generator to construct effective reasoning trajectories more efficiently, and improves the accuracy and verifiability of generation.
    \item \textbf{AST simplification and transformation}: Rule-based autoformalization methods often create syntax rules, generate corresponding AST trees, and then convert them to formal language through linearization methods.
\end{itemize}
    Table \ref{tab:your-table} provides a comprehensive overview of the datasets used for theorem proving and automatic formalization, including the addresses, sizes, and brief descriptions of the datasets.
    Table \ref{tab:your-table2} gives the support that the individual datasets contain for the general dataset elements. Includes informal problem statement, formal problem statement, informal solution, formal solution.
    The main task is to summarize the task categories supported by the datasets.
    Each element in Table 3 is described as follows:
    \begin{itemize}
        \item \textbf{Informal Statement}:It refers to the description of mathematical problems expressed in natural language, usually close to the way of expression in mathematics textbooks or Olympiad questions, suitable for human understanding but not strictly following the grammar of formal language.
        \item \textbf{Formal Statement}:Problem definitions accurately represented by formal languages (such as the syntax of proof assistants like Lean, Isabelle, and Coq) possess logical precision and verifiability, and can serve as objective theorems in formal systems.
        \item \textbf{Informal Solution}:The problem-solving process or answer expressed in natural language or standard mathematical writing forms usually includes reasoning steps, auxiliary lemmas, etc., which are convenient for manual review but do not have verifiability in a formal system.
        \item \textbf{Formal Solution}:The corresponding formal language code, which is usually checked by a theorem proving assistant, is the core component to achieve automatic verification and program aided proof.
    \end{itemize}

\section{Workflow for Autoformalization and Theorem proof}
\label{Work Flow}
Based on the solid foundation laid by the open source method models organized in Table\ref{tab:my-table} and the latest research progress in the field of autoformalization, we turn our perspective to the key dimension of workflow, and deeply analyze the mechanisms and concerns of autoformalization in different links, aiming to provide an overall framework as shown as Figure \ref{fig:enter2-label} to comprehensively reveal the workflow of automformalization.
\subsection{Data preprocessing}

The NuminaMath team notes that ``Good data is all you need''\cite{numina} is important for Autoformalization because data is more scarce than other tasks. Therefore, the requirements for data quality and quantity are higher, and the data preprocessing part is divided into two methods: enhancing data quality and increasing data quantity.

\subsubsection{Enhanced data quality}

First, you can enhance data quality by filtering low-quality data. \cite{xin2024advancing} presents a novel approach to generating valuable Lean4 proof data from natural language math problems at the high school and undergraduate levels, and filtering low-quality data through model scoring and hypothesis rejection methods. The model score, in which DeepSeek-Prover rates the generated formal statements, was classified as ``excellent'', ``good'', ``above average'', ``average'', or ``poor''. Statements that are rated ``fair'' or ``poor'' will be filtered. Hypothesis rejection, that is, let DeepSeek-Prover try to prove false propositions, if successful proves it is filtered. This also reminds us that filtering low-quality data can limit the data from multiple dimensions, and make the informal statement as much as possible logically consistent and semantically consistent with the corresponding formal statement. The article uses the logic of true and false propositions, but also can be other such as theory resolution. Perhaps adding variety to each area of knowledge would improve the logic of autoformalization.

\subsubsection{Increase data quantity}

Since the data of Autoformalization is scarce, increasing the amount of data in the data preprocessing stage can also help improve the performance of the model. Specifically, this can be done by adding a variety of formal languages \cite{NEURIPS2024_984de836}; Expand the search scope \cite {wu2024leangithubcompilinggithublean}; Synthetic data \cite {azerbayev2023proofnetautoformalizingformallyproving}, such as the above method to increase the amount of data.

Data from multiple formal languages facilitates autoformalization. \cite{NEURIPS2024_984de836} translated two of the largest corpora: Archive of Formal Proofs (Isabelle) and mathlib4 (Lean4) into natural language using GPT-4. This process is mainly based on a key observation: informalization is much simpler than formalization, that is, the quality of data translated from formal languages to informal languages is higher than that from informal languages to formal languages, and powerful LLMs can produce different natural language outputs. Through ablation experiments, data from multiple formal languages is essential for autoformalization training.

Due to a large number of written by human still didn't get the full use of formal language corpus, in order to solve this problem \cite {wu2024leangithubcompilinggithublean} collects almost all of the Lean4 repositories on GitHub, and makes suitable training data; From Mathlib4's 10,000 historical real-world commits, \cite{xin2025apebenchifilelevelautomated} builds the first large-scale, file-level dataset. Both of these two methods increase the amount of data, and experiments show that training on such a large scale dataset, the model improves performance in a variety of mathematical problem domains.

In order to further expand the data, \cite{liu2025atlasautoformalizingtheoremslifting} extracts mathematical concepts from mathlib, and randomly selects two concepts to let the LLM generates informal problem statements, and uses the model with more parameters as the teacher model to generate the corresponding formal proofs, thus creating a new massively parallel corpus. In order to further expand the data, \cite{liu2025atlasautoformalizingtheoremslifting} also disproves the assumptions in the formal proof, so that the number of parallel corpora created in the end is nearly doubled. Using the final created parallel corpus, the student model is guided through reinforcement learning methods to iterate over its autoformalization capabilities.

\subsection{Method}

\subsubsection{Rule-based Autoformalization}

Rather than relying on data-driven machine learning models, the core is to leverage explicit logical, syntactic, and semantic rules to enable structural transformations. To address the flexibility and complexity of natural language many systems have adopted controlled natural language - allowing users to express mathematical proofs in a way that is both natural and formal, giving rise to systems including Mizar, NaProche, ForTheL, MathNat, and VerboseLean. At the same time, attention has been given to the GF advanced grammar writing tool, which allows the flexible development of custom grammas to parse mathematical texts directly. Representative works of Autoformalization based on GF: \cite{pathak2024gfleanautoformalisationframeworklean} proposed a framework can use GF to translate natural language into formal language, the algorithm is implemented in Haskell to build an abstract syntax tree for simplified ForTheL expressions, and a Lean expression syntax is implemented in GF. The Lean expression can be obtained by linearizing the abstract syntax tree obtained in the previous step to obtain a framework for symbolic natural language understanding and processing: GLIF.

\subsubsection{LLM-based Autoformalization}

Unlike rule-based methods, LLM-based methods are more flexible and can capture details in natural language that experts may miss when creating rules. LLMs like GPT4 represent a new paradigm for machine learning. Such autoregressive models have been pretrained using large amounts of Internet data and can quickly perform a variety of downstream tasks by leveraging few-shot learning capabilities. Recent research has shown that LLMs can translate between informal and formal mathematical statements with less than 5 examples, suggesting that we may not need to collect a large-scale formal corpus to achieve autoformalization , although large-scale training data is still an issue of concern when training models. It is also worth noting that informalization is usually easier than formalization. Under the same model, the informal math competition level problem achieves 70$\%$ accuracy, while the formal math competition level problem only achieves 30$\%$ accuracy. Exploiting this observation has also facilitated subsequent research on LLM-based back-translation. 

There are some key differences between informal languages and formal languages: 1) Semantic gap: informal languages are polysemy in different contexts, while formal languages require a single definition without ambiguity; 2) Logic gap: informal languages often omit tacit knowledge, while formal languages need to explicitly fill in the completeness, that is, theorems that may be intuitive in formal languages also need to be split into atomic steps. \cite {JiangWZL0LJLW23} guides the automatic prover to complete the autoformalization task by mapping the informal proof to the formal proof step. The purpose of this guidance is filling the semantic gap between the informal language and the formal language, and establish a more complete logic chain indirectly. In order to study how code training affects mathematical reasoning, \cite{xin2024advancing} conducted ablation experiments. The results show that two-stage training of code training and mathematical reasoning is better than simultaneous training of code training and mathematical reasoning, and the latter is even worse than direct training of mathematical reasoning. It speculated that DeepSeek-LLM 1.3B was small and lacked the ability to fully assimilate code and mathematical data at the same time, which was actually more logical. After all, the gap between informal languages and formal languages was large, and this gap could not be directly and completely filled, so various methods were derived to supplement the gap between them.

In natural language machine translation related work, the quality of translation largely depends on the quality of parallel data between the two languages. But such parallel data are rare and difficult to manage, and back-translation is one of the most effective ways to improve the quality of translation. \cite {azerbayev2023proofnetautoformalizingformallyproving} uses OpenAI distillation of the Codex model of back translation, in order to improve the model of automatic formal ability. \cite{NEURIPS2024_984de836} They used GPT-4 to generate informal language data from an existing library of formal languages. They have empirically and analytically concluded that informal is much simpler than formal, so back-translation is part of the Autoformalization effort.\cite{chan2025leaningqualityhighqualitydata} explores three approaches to back-translation and constructs a high-quality, small-scale dataset, which outperforms MMA,ProofNet using only 1/150 of the number of tokens. First, it trains a single model to handle both directions of the translation task, it does not employ a teacher model and a student model as in the traditional practice: (1) Transform a batch of examples to informal language; (2) Using these synthetic informal language examples and formal language example pairs as labeled training data, the synthetic informal language examples are back-translated to formal language; (3)Compute the loss between the generated formal language examples and the original formal language examples; (4)Update the model weights via backpropagation. Experiments significantly reduce the evaluation loss after only a few hundred training steps, resulting in a significant improvement over the baseline; Another method uses a pretrained model with more parameters and stronger performance as a teacher model, provides an formal language example dataset, and uses the model to generate the corresponding synthetic informal language text, and then trains the student model to generate formal language according to the informal language, so that the ability of the small-scale student model approaches that of the large-scale teacher model.The last method captures the specific keyword parsing formal proof strategy through regular expressions, which is more interpretable than the above two methods, but due to the limitations of the format, only partially simple translations can be generated, and it is difficult to surpass the above complex back-translation methods.

\subsubsection{Postprocessing}

In the post-processing stage, it is generally ensured that the formal output corresponds to the informal input by modifying the mechanism or solving the semantic illusion of the statement and the logic error of the proof in Autoformalization. The semantic illusion of statements refers to the mismatch between the generated formal language statements and the natural language statements. \cite{xin2024advancing} suggest a hypothesis rejection method which is used to filter out the semantic illusion caused by the partial self-contradiction of statement hypotheses after autoformalization. The principle that contradiction implies false proposition is used to filter out the self-contradictory false proposition. \cite {lu2024formalalignautomatedalignmentevaluation} is the first automated assessment alignment method -- by minimizing the contrastive loss make cosine similarity of corresponding informal - formal pair is higher than not . In this way, the alignment is correctly evaluated and the semantic illusion is alleviated. The logic error of proof mainly refers to the break of proof logic caused by using wrong proof tactics when it is converted into formal language proof. \cite {zheng2024lyraorchestratingdualcorrection} suggests tool correction and guess correction mechanism is put forward, including tool correction mechanism to modify LLM formal certificate used in the error generated by the strategy, avoid because of the large model error using proven strategies lead to prove logic fault.

\subsubsection{Evaluatation}

In the Autoformalization workflow, syntactic accuracy and semantic equivalence at the evaluation level are key factors to ensure the success of the autoformalization process. Syntactic accuracy refers to the ability to correctly translate into formal language, that is, whether the final formal result can pass the type check of Lean, Coq and other environments. It is also possible to extend first-order logic formal languages with syntax close to natural languages but more mathematically expressive. \cite{zhou2025stepwiseformalverificationllmbased} uses a LLM to translate the natural language problem solution steps into a kind of designed formal language SimpleMath, and take a data-oriented approach to structure the formalized solution into a solution Graph, which shows the dependencies of each reasoning step. When verifying the correctness of a reasoning step, it does not need to input all constraints into the verification model, but selects the most relevant premise to the current reasoning conclusion, which greatly reduces the input length and improves the efficiency. Compared with formal languages such as Lean and Coq, it is more flexible. The type checking of Lean, Coq and other environments does not mean that it is completely correct, because the formalized results also need to be semantically equivalent to the corresponding natural language version, so it is equally important to evaluate the semantic equivalence. The purpose of grammatical accuracy is to ensure the result of autoformalization through the formal language proof assistant's type checking mechanism, which mainly involves two aspects: 1) at the level of code generation: the generated formal language code has no grammatical errors; 2) At the level of mathematical reasoning: the generated formal language code has a correct logical chain. The most common way to assess accuracy is to compare automatically generated formal expressions with the results of expert manual formalization. Such evaluation often rely on recall metrics or manual inspections(\cite{wu2022autoformalizationlargelanguagemodels} \cite{agrawal2022mathematicsformalisationassistantusing}, \cite{azerbayev2023proofnetautoformalizingformallyproving}, \cite{jiang2023multilingualmathematicalautoformalization})to judge the consistency of the two in mathematical content. Although it is right, this method is time-consuming and subjective, so it is commonly used to calibrate automatic evaluate tools. For the accuracy of mathematical reasoning, such as whether the model corresponds to the specific meaning of each variable, or whether the proof strategy used in the proof process is correct, For the former, \cite{li2024autoformalizemathematicalstatementssymbolic}suggests the method of symbolic equivalence . Split statements into premises and conclusions (assuming that premises do not contradict themselves) When the premises and conclusions of the formal results to be verified are respectively equivalent to the corresponding correct premises and conclusions by ATP proof, it is explained that the model corresponds to the specific meaning of each variable. For the latter \cite{zheng2024lyraorchestratingdualcorrection} leverage the feedback to evaluate whether the proof strategy is correct and the tool correction mechanism is used to resolve it. For semantic equivalence, the main judgment is whether the formal expression generated by autoformalization accurately reflects the corresponding natural language expression. Relevant indicators commonly used similarity measures including text similarity (e.g., BLEU, ROUGE, etc.), \cite{poiroux2025improvingautoformalizationusingtype} presents the symbolic computation index: BEqL and BEq +. As pointed out by \cite{liu2025rethinking}, BEqL can indicate the accuracy of various syntactic changes. This method can achieve the accuracy of various syntactic changes. Although it has a high false negative rate, the paper believes that its simplicity makes this indicator worth using. BEq+ introduces more complex policy combinations (such as simp, ring, etc.) on the basis of BEqL, further enhancing the ability to verify the accuracy of deep mathematical structures (such as algebraic identities). \cite {lu2024formalalignautomatedalignmentevaluation} mentioned that for LLM generated formal results, even if the BLEU index is high also don't tell the formal result with the corresponding natural language expression is semantic equivalence, Alignment scoring is also proposed to better evaluate semantic equivalence.      Recent studies have also proposed non-numerical metrics to verify semantic    equivalence: \cite{xuejun2025mathesisformaltheoremproving} proposes the LeanScorer method, which uses a LLM to decompose natural language statements into premises and conclusions. Then, another LLM is used to score the premises and conclusions corresponding to the formal statements generated by the model respectively. Finally, the scoring results are integrated into numerical scores to test the semantic equivalence of the formal statements;\cite{raza2025instantiationbasedformalizationlogicalreasoning} proposes a new semantic verification method, semantic self-verification, which can solve the key challenge of combining language models with the rigor of logic solvers without training a model: how to more accurately translate reasoning problems in natural language into a formal language for solvers. \cite{raza2025instantiationbasedformalizationlogicalreasoning} generates the solution program through LLM, and then generates specific positive and negative examples to verify whether the formal results meet the two kinds of test instances to judge the semantic correctness of the formal results. It also uses examples to constrain, so that LLM can better modify the failed formal results.

\subsection{Therom Proof}
While autoformalization emphasizes the transformation of informal statements into formal representations, its practical utility is ultimately grounded in the ability to construct complete and verifiable proofs.  As such, recent advances in large language model–driven theorem proving not only complement autoformalization but also expand its scope—by generating formal derivations that validate the translated statements. \cite{azerbayev2024llemmaopenlanguagemodel} suggests LLEMMA, 7B and 34B parametric language models specifically designed for mathematics, which leverage the few-shot learning capabilities of LLM to achieve good performance on several benchmarks. \cite{xin2024advancing} released the base model DeepSeekMath with 7B parameter sizes achieving comparable performance to Minerva540B, which also shows that the number of parameters is not the only key factor in mathematical reasoning ability, and strong performance can also be obtained in smaller models. These methods can be broadly categorized by their granularity: some focus on the step-by-step generation of tactics, enabling precise control and local reasoning, while others aim for holistic, end-to-end proof generation.  We begin by reviewing the former, which centers around tactic-level reasoning and search, and plays a foundational role in training models to navigate the formal proof space effectively.

\subsubsection{Tactic Step Proving Methods}
Recently, many works have been devoted to enhancing the local reasoning ability of automated theorem proving using tactic steps. For example, \cite{xin2025bfsproverscalablebestfirsttree} proposes a scalable expert iteration framework based on breadth-first tree search. In each iteration, the framework filters the question set, and combines Direct Preference Optimization and length normalization strategies to optimize the language model. It makes it preferentially explore promising proof paths. Bfs-prover achieves a 72.95\% pass rate on the Lean4 MiniF2F benchmark, showing that simple BFS search can be competitive under appropriately scaled conditions. 

Similarly, \cite{wu2024internlm25stepproveradvancingautomatedtheorem} uses an expert iteration strategy on large-scale Lean math problem sets. We also train a critic model to pick relatively simple exercises to guide the model to search for deeper proofs. By optimizing question selection and deepening the exploration depth, our method achieves 65\% pass rate on MiniF2F and proves 17.0\% of questions on the Lean-Workbook-Plus dataset. These methods mainly focus on strengthening local search strategies, improving model sample efficiency, and laying the foundation for more complex global proof generation.

\cite{li2025hunyuanproverscalabledatasynthesis} aims to alleviate the problem of training data sparsity: it continuously generates new training samples through a scalable data synthesis framework, and designs a guided tree search algorithm to support efficient inference. Through this combination of data iteration and search method, \cite{li2025hunyuanproverscalabledatasynthesis} achieves 68.4\% pass rate on benchmarks such as MiniF2F. Furthermore, \cite{li2025hunyuanproverscalabledatasynthesis} has published a dataset containing 30,000 synthetic proof instances, each of which contains a natural language question, an automatically formalized Lean statement, and the corresponding proof. The above tactical step proof method not only optimizes the local search strategy and improves the proof performance, but also provides valuable data support for autoformalization by synthesizing natural language and formal proof pairs.

\subsubsection{Whole-Proof Generation Methods}
End-to-end proof generation methods have also made significant breakthroughs recently, especially in integrating natural language reasoning and formal proof.

\cite{lin2025goedelproverfrontiermodelopensource} adopted another end-to-end training idea. Firstly, large-scale LLM was used to automatically translate natural language math problems into Lean 4 formal statements, and the Goedel-Pset-v1 dataset containing 1.64M formulative statements was generated. We then continuously train a series of Prover models so that the new model can prove statements that the old model failed to solve, continuously expanding the training set. Based on this dataset, \cite{xin2024deepseekproverv15harnessingproofassistant} is supervised and fine-tuning, and the resulting Goedel-Prover-SFT achieves a success rate of 57.6\% on MiniF2F, which is 7.6\% higher than the previous optimal model. This approach fully demonstrates the effectiveness of automatically generating large-scale training data and iteratively optimizing the model.

Based on the Qwen2.5-72B model, \cite{wang2025kiminaproverpreviewlargeformal} proposes a structured "formal reasoning model" combined with a large-scale reinforcement learning pipeline to imitate human problem-solving strategies. The model achieves a pass rate of 80.7\% on the MiniF2F benchmark, and has demonstrated its powerful ability under very low sampling conditions, showing its high sampling efficiency. \cite{wang2025kiminaproverpreviewlargeformal} also demonstrates a clear positive correlation between model size and performance, and its unique style of reasoning also potentially bridges the gap between formal verification and informal mathematics.

\cite{zhang2025leanabellproverposttrainingscalingformal} focuses on post-training extension of existing proof models: it first continuously trains the base model on a mixed dataset that includes a large number of formal statement-proof pairs as well as auxiliary data that mimics human reasoning behavior, and then leverages Lean 4 compiler feedback as a reward signal for reinforcement learning to optimize the model. Through this combination of continuous training and reinforcement learning, \cite{zhang2025leanabellproverposttrainingscalingformal} successfully improves the ability of models such as \cite{xin2024deepseekproverv15harnessingproofassistant} and \cite{lin2025goedelproverfrontiermodelopensource}, making it reach a pass rate of 59.8\% on MiniF2F.

\cite{ren2025deepseekproverv2advancingformalmathematical} introduces reinforcement learning and subgoal decomposition techniques: In this method, the existing model DeepSeek V3 is first used to decompose the complex problem into a series of subgoals, and then the proof process of the solved subgoals is combined with the stepwise reasoning of the original model to form a complete solution chain for cold start reinforcement learning training. This strategy of integrating natural language reasoning and formal proof enables the \cite{ren2025deepseekproverv2advancingformalmathematical} model to achieve a pass rate of 88.9\%on the MiniF2F test set, and solve 49 problems on PutnamBench.

In summary, the above end-to-end approach not only greatly improves the automatic proof capability, but also Bridges the gap between natural mathematical languages and formal mathematical proofs. These advances provide solid support for autoformalization research more training data and stronger proof power are pushing autoformalization forward.

\section{Taxonomy of Autoformalization for Mathematical Problems}
Mathematics, as a rigorous and systematic discipline, its autoformalization has unique and significant meaning and is also the basis for exploring the completely reliable reasoning of LLM.   The autoformalization of mathematics shows broad application 
prospects and theoretical value in multiple dimensions.   Next, we will conduct a detailed analysis of the autoformalization in the field of mathematics from multiple perspectives, includes the domain of mathematical problem, the difficulty of the mathematical problem, and the analysis of the mathematical problem in terms of the level of abstraction.

\subsection{Domain of Mathematical Problem}
\label{Domain division of mathematical problems}
To better illustrate the diversity and challenges of autoformalization across mathematical domains, we now delve into the representative subfields where recent progress has been most notable. Each domain has its own unique characteristics such as graphical reasoning in geometry, symbolic manipulation in algebra, and different requirements put forward different requirements for formal systems. A detailed analysis of these areas leads to a clearer understanding of how autoformalization must adapt to different semantic structures, input modalities, and reasoning paradigms.

\subsubsection{Geometry and Topology Problems}

The characteristic of this type of problem is that a large part of the information in the description of the problem is contained in the graph. The core difference from other problem domains is information complexity: geometry and topology problems deal with both natural language descriptions and graphical elements (e.g., geometry, coordinate systems, topological relationships). For example, graphics in geometry problems have hidden attributes such as angle and side length, and graphic information needs to be encoded into symbolic logic (such as through coordinate systems or geometric theorems). The difficulty lies in feature extraction of graphics and the need to align the semantics of text and graphics (such as vertex position matching of triangle A, B, C), while most other problem areas only rely on natural language descriptions. \cite {murphy2024autoformalizingeuclideangeometry} solves the challenge in the field of automatic formal about Euclidean geometry: the informal proofs rely on chart, leaving the defects of difficult to formalize. Implicit graphical reasoning is common in informal geometric proof, such as the proof uses the intersection of two circles in the graph, but does not prove their existence, so the model is needed to fill the defects of graphical reasoning. To fill this defect, \cite {murphy2024autoformalizingeuclideangeometry} using a formalized system called E, the system of extracting a set of graphical rules, so can be modeled as a logic diagram reasoning, in order to formalize.

\subsubsection{Algebra and Number Theory Problems}

This kind of problem is characterized by symbolic logic. The core of Autoformalization of this kind of problem is to convert variables, equations and operation rules described by natural language into symbolic logic chains. For example, linear algebra problems (such as matrix operations) require mapping concepts such as vector Spaces described in natural languages to formal languages such as vectorspaces, where the difficulty lies in resolving nested symbolic structures and the priority of implicit operations. Since mathematical languages contain features that are not common or exist in natural languages, such as numbers, variables, and defined terms, \cite{DBLP:journals/corr/abs-2301-02195} proposes a semantic parsing method based on Universal Transformer, This includes a replication mechanism that introduces a different generic tag for each unique numeric constant or variable literal in the theorem and its proof, and by using generic tags for numbers, variables, and definitions, with only a limited set of embeddings to train, the model is able to generalize to invisible numbers or variables and define names, and to some extent support out-of-vocabulary mathematical terms. Compared with geometry and topology problems, algebraic and number theory problems focus more on references and nested relationships between various variables, as well as implicit operational logic. This is the comfort zone of LLM, using LLM's powerful logical reasoning ability to deal with plain text, you can better solve the problem without cross-modal alignment.

\subsubsection{Complex Systems and Frontier Problems}

Such problems often directly address practical challenges, bridging theoretical and real-world needs, and include cutting edge mathematics that has not yet been fully formalized. Examples include partial differential equations, which are used to describe complex systems that interact across time and space. For example, the system applied to the rod by heat and mechanical stretch is modeled and controlled by heat and wave equations. \cite{soroco2025pdecontrollerllmsautoformalizationreasoning} uses signal temporal logic to formalize the constraints in the partial differential equation control problem, so as to clearly and accurately express the complex specification constraints into logical formulas and eliminate possible ambiguity. And then leverage LLMs to complete Autoformalization task. It has great potential for such complex system problems in promoting scientific and engineering applications. Frontier Math covers hundreds of undisclosed problems, preventing large models from "memory the data" to solve them. The powerful GPT-O3 only solved 25.2$\%$problems, while other LLMs only solved 2$\%$problems. It reveals that there is still a large gap between the existing large models and human mathematical experts in the dimension of creative reasoning, which is also a problem faced by the current Autoformalization field.

\subsection{Difficulty Level of Mathematical Problems}
\label{Difficulty level}

Autoformalization can also be divided into three levels of difficulty according to the difficulty level of the corpus: high school, undergraduate and graduate mathematics problems. Different levels of mathematical problems are not only different in language expression, but also in the complexity of the proof: for example, the length of the proof, the type and number of strategies used, and the level of abstraction show a step increase.

\subsubsection{High School Level}

High school math problems mainly focus on high school math competition problems, such as IMO, AIMC and other math competitions. Problems usually have a clear idea, a short proof process, and a standardized logical structure, and generally involve basic algebraic, geometric, and preliminary number theory propositions. \citeauthor{zheng2022minif2fcrosssystembenchmarkformal} proposes a representative data set miniF2F problems are mostly taken from high school Olympiad math problems, with clear structure and limited steps.

\subsubsection{Undergraduate Level}

Undergraduate-level math problems are mostly from classical math textbooks, covering the fields of real analysis, linear algebra, abstract algebra, topology, etc. The representative data set is ProofNet presented in \cite{azerbayev2023proofnetautoformalizingformallyproving}, which contains 371 examples, each of which is composed of formal theorems in Lean 3, natural language theorem statements and proofs. It covers real analysis, complex analysis, linear algebra, abstract algebra and topology. \cite{yu2025formalmathbenchmarkingformalmathematical} and \cite{liu2025combibenchbenchmarkingllmcapability} include theorems from both high school Olympic competitions and undergraduate levels. \cite{yu2025formalmathbenchmarkingformalmathematical} covers a variety of fields, while \cite{liu2025combibenchbenchmarkingllmcapability} and \cite{xiong2025combinatorialidentitiesbenchmarktheorem} both focus on building benchmarks for combinatorial math problems. The FormL4 dataset presented in \cite{lu2024processdrivenautoformalizationlean4} extracts theorems from Mathlib4 and generates corresponding informal descriptions through a large language model, reflecting the step-by-step progression of mathematical theorems from foundation to complex proofs in university. The problem involves a higher level of abstraction, the proof process is long and rigorous, and the model needs to understand high-level mathematical concepts.

\subsubsection{Graduate Level}

Problems at the graduate level and beyond are usually derived from the latest academic papers or advanced mathematical monographs, and involve a higher degree of theorems background, proof strategies, and conceptual abstraction than at the undergraduate level. \citeauthor{patel2024newapproachautoformalization} proposes a representative dataset named arXiv2Formal, sampled from mathematical papers on arXiv, and contains 50 research-level theorems, focusing on ``unlinked formalization'', i.e. when the necessary background definition is missing in the formalization process, and the model needs to be ``unlinked formalization'', i.e. in the absence of a full context, the model needs to be ``unlinked formalization''. Complete the connection between definitions, lemmas and theorems. Models need to deal with highly abstract mathematical concepts, understand complex proof strategies, and reason and link in the absence of full context; The Herald dataset \cite{gao2025heraldnaturallanguageannotated} provides natural language notes and formal proof pairs generated by Mathlib4 in Lean 4, aiming to provide fine-grained corpus of advanced mathematical proofs for automatic formal systems. It is mainly for graduate students and above in mathematical research, and contains high-level mathematical languages and proof strategies. It is suitable for training models to deal with complex mathematical proof tasks. 

\subsection{Abstraction Level of Mathematical Problems}
The proof length and the number of tactics used provide a visible indication of difficulty across the datasets for the three educational levels discussed above. Most problems align well with the proposed difficulty classification. This section further examines the \textbf{level of abstraction} as a complementary dimension for differentiating problem difficulty.
Many mathematicians view abstraction as a form of generalization. For example, \citeauthor{abstraction} defines abstraction as follows: *The abstraction of a concept $C$ is a concept $C'$ that includes all instances of $C$, constructed by adopting as axioms those statements that are universally true for all instances of $C$. A classical example is the notion of a \textit{group}, which abstracts the set of all symmetries of an object. When binary operations are interpreted as symmetric compositions, the group axioms capture the shared structure of all such transformations. For instance, the symmetries of regular polygons, such as rotations and reflections, can be combined to form a set that satisfies the group axioms. Thus, the idea of a group arises by abstracting these shared regularities into a general concept.
To further illustrate how abstraction level evolves across educational stages, we analyze how the concept of ``function'' is treated in high school, university, and graduate-level mathematics problems.

\subsubsection{Procedural Abstraction in High School Mathematics}

According to \cite{AdvancedMathematicalThinking}, the understanding of functions at the high school level typically remains at the stage of \textbf{procedural abstraction}—characterized by reliance on techniques rather than formal axiomatic reasoning. High school mathematics is relatively low in abstraction, with content tightly organized around concrete examples and intuitive interpretations.
For example, finding the zeros of a quadratic function usually involves applying procedural techniques such as computing the discriminant or analyzing the graph of the function. Functional properties like monotonicity or evenness are verified through direct algebraic manipulation, with strong dependence on numerical values rather than logical formalism. For instance, to demonstrate the monotonicity of $f(x) = e^x$, one computes its derivative—a method grounded in concrete computation rather than abstract reasoning.

\subsubsection{Logical Abstraction in University Mathematics}
While abstraction is present in all stages of mathematical thinking, university-level problems typically require a higher degree of \textbf{conceptual and logical abstraction}. A case in point is the concept of the limit. In high school, limits are often introduced through numerical approximation, whereas university mathematics formalizes them using symbolic logic, completely decoupling the concept from concrete numerical values.
This shift reflects a fundamental change: mathematical objects are now approached through logical relationships, and specific contextual examples are abstracted away. The focus is on rigorous definitions and logical derivations, which demand a significantly deeper cognitive engagement than the procedural reasoning typical of high school problems.

\subsubsection{Structural Abstraction in Graduate Mathematics}
At the graduate level, mathematical problems attain an even higher level of abstraction—one that emphasizes the \textbf{structure of mathematical spaces}. Functions are no longer treated merely as computational objects or graphable expressions; they are viewed as elements of abstract spaces defined by properties they satisfy.
For example, the $(L^p)$ space consists of $p$-integrable functions, while the $\ell^p$ space consists of $p$-summable sequences. The mathematical focus shifts from analyzing specific functions to understanding the properties of the spaces they inhabit—such as completeness, compactness, or duality.
In this stage, solving a problem often involves reasoning about the internal structure of these function spaces, constructing mappings between them, or proving structural properties. This level of abstraction demands a highly sophisticated understanding of mathematical systems and their interrelations.

\section{Connecting LLM Outputs and Formal Verification via Autoformalization}

Autoformalization plays a crucial role in applying formal verification methods to large language model (LLM) outputs. Despite their powerful capabilities, LLMs often suffer from significant hallucinations, undermining the reliability and trustworthiness of their generated results. Consequently, verifying the correctness and trustworthiness of LLM-generated outputs has become a pressing research topic. Achieving fully trustworthy outputs requires rigorous verification of the underlying reasoning processes, which essentially translates into mathematical and logical verification tasks. Such tasks can leverage various formal verification frameworks, including first-order logic, higher-order logic, and formal specification languages. Regardless of the chosen formal verification approach, a common challenge is accurately translating—or autoformalizing—the informal outputs produced by LLMs into formal representations suitable for verification. Autoformalization thus serves as a critical bridge connecting LLM reasoning with formal verification techniques. 

\subsection{Code validation}
In the field of formal verification of LLMs output, the traditional research content is code verification. Whether the code generated by the large model is correct directly affects whether the code can run. In the field of code generation, the cost of software failures is estimated to be in the trillions of dollars per year in the US alone \cite{report}. Since the advent of LLMs, people have placed their hopes on the powerful ability of LLMs to generate code. However, the verifiability of the generated code is poor at present, so it is crucial to ensure the correctness of the generated code.

\subsubsection{Non-interactive validation}
Static analysis is a technique to analyze program code without executing the program and interacting with other verifiers. It aims to detect errors in the program, verify the properties of the program, or extract structural information of the program. Static analysis plays an important role in formal verification. It provides a systematic way to ensure the correctness and soundness of programs.

\textbf{Dependency graph}
\cite{liao2025augmenting} designs and builds dependency graphs based on the results of static analysis, aiming to capture key semantic information in decompiled code to verify semantic consistency. The dependency graph is a triple DG= (V,E,L), where V represents the set of nodes and contains all variables and related expressions in the decompiled code. E represents the set of edges, including control flow dependent edges, state dependent edges and type dependent edges. L is the labeling function used to map edges into the three dependencies.

\textbf{Assertion generation}
In the domain of code verification, assertion generation is a key method of automated testing, used to ensure that code units behave as expected. This process mainly revolves around the writing and execution of test cases. Test cases usually consist of two parts: test prefixes (the code that sets up the test environment and calls the function under test) and assertions (used to verify whether the function output or program state is correct). The goal of assertion generation is to automatically generate assertion code to reduce the effort of manually writing test cases and improve test efficiency. In code verification, this step is crucial because it directly determines the validity and accuracy of the test cases.

\textbf{Specification generation}
Specifications are often used in code validation to describe the expected behavior of a function, module, or entire system, ensuring that the developer understands the requirements and implements the functionality as expected. \cite{ma2024specgen} effectively addresses the shortcomings of traditional approaches through two phases: dialogue driven and mutation driven. The dialog-driven phase improves the quality and accuracy of the specification by using the feedback information from the verifier to gradually guide the model to generate accurate specifications through multiple rounds of interaction with the LLM. In the mutation-driven phase, for the failed specifications generated by LLM, the mutation operators are applied to modify them, and the most likely specifications are screened out through the heuristic selection strategy, so as to further improve the success rate of specification generation.

\textbf{Loop invariant generation}
Loop invariant generation describes the properties that a loop must satisfy at each iteration. The goal is to find a set of logical propositions for a loop in a program that holds at each iteration of the loop. These invariants are crucial for formal verification because they allow verification tools to locally check the correctness of loops without considering the rest of the program. \cite{kamath2023finding} presents Loopy, which consists of two main phases. First, LLMs is used to generate candidate loop invariants. Second, these candidate invariants are verified to be correct using a symbolic tool such as Frama-C. If the invariants generated by LLMs are incorrect, Loopy uses the repair process to adjust these invariants so that they are correct. This method effectively utilizes the powerful generation ability of LLMs and the precise verification ability of symbolic tools.

\subsubsection{LLM interactive verification}
The combination of code and LLMs is verified by using an interactive theorem prover based on first-order/higher-order logic and type systems. LLMs can use the feedback from formal validators to filter out the correct programs.

\textbf{LLM generation}
\cite{aggarwal2024alphaverus} uses a large language model to translate the high-resource Dafny code into the low-resource Verus code, solve the problem of data scarcity, and generate candidate programs. Then, the generated code is optimized by using the tree search algorithm and the feedback interaction information of the validator to make it correct.

\textbf{LLM judgment}
\cite{liao2025augmenting} uses LLM to infer the correct type of variables through static analysis of the supplemented context information and guides the optimization process through reasoning chain prompts. SmartHalo will convert the dependencies derived in static analysis into natural language descriptions (such as "The type of variable A depends on expression B") and input them as prompts to the LLM. These prompts help LLMS understand the static constraints of the optimization objectives, thereby making more accurate predictions.

\subsection{Medicine Domain}
 In the medical field, the application of large language models (LLMs) holds great potential and can provide support for medical diagnosis, drug development, and patient care, among others. However, due to the particularity of the medical field, it has extremely high requirements for reliability and accuracy. Medical decisions are often directly related to the safety and health of patients. Therefore, it has become a crucial task to ensure the accuracy and reliability of the system output based on LLMs by using formal verification. \cite{hamed2025knowledge} employs a non-interactive verification method, using simple binary relation association verification to verify the reliability of the associations generated by LLMS. This formal representation provides a structured approach to model the associations between different categories of terms. Verify the semantic accuracy and reliability of the terms generated by the large model in the disease, gene, and drug ontology datasets. Correctness verification: Use biomedical ontologies such as GO, DOID, ChEBI, and symptom ontologies as true criteria to verify the identity of terms. When a certain term is not found in the ontology, it is called "unverified". Reliability verification: Represented by binary relations. For example, the association between disease and gene can be expressed as RDG $\subseteq$ D×G, where D is the set of disease terms and G is the set of gene terms. A pair (d, g)$\in$ RDG indicates that a specific gene g is associated with a specific disease d. This formal representation provides a structured approach to model the correlations among terms of different categories. 

 \subsection{LLM output verification}
 \citeauthor{song2025mindgapexaminingselfimprovement} explores a framework in which a model validates its own output. Based on this verification, filter or reweight the data and extract the filtered data. And a comprehensive, modular and controlled LLM self-improvement study was initiated. This article provides a mathematical formula for self-improvement, mainly controlled by a formal generative validation gap (GV-Gap). The SymGen method \cite{hennigen2023towards} uses ``symbolic references'' to establish associations between natural language texts and structured data. Symbolic references provide clear data source references for autoformalization by embedding reference identifiers for specific data fields (such as \{filename\}) in the text. This reference mechanism is similar to model annotation or marking in formal methods, which can help formal tools quickly locate the data elements on which the text content depends, thereby achieving more efficient data-driven formal transformation. It not only helps ensure the accuracy and traceability of the content in the text generation stage, but also plays a key role in the subsequent formal verification and maintenance process.

\section{Open Challenges and Future Dirctions}

While autoformalization has made promising strides in recent years, several critical challenges must be addressed to unlock its full potential—particularly as we move toward large-scale, trustworthy AI systems for mathematical reasoning.

One of the most pressing issues is data scarcity. Although large language models (LLMs) have benefited from vast amounts of open-domain textual data, autoformalization requires high-quality parallel corpora consisting of aligned informal and formal mathematical statements. Due to the complexity and expertise required to produce formal proofs, such data remains limited in both scale and diversity. Current resources—such as the LeanWorkbook dataset (4.8MB)—represent important progress, yet are small compared to the massive datasets used to train LLMs in other domains. Addressing this gap will require the development of more scalable data collection pipelines, improved techniques for synthetic data generation, and perhaps human-in-the-loop systems that incrementally refine formal annotations with minimal expert intervention.

Looking forward, the ability to formalize larger and more abstract theories—without the need for extensive fine-tuning—will be crucial. As highlighted by \cite{wu2022autoformalizationlargelanguagemodels}, relying on full retraining is computationally expensive and impractical at scale. Instead, we envision autoformalization systems that can generalize compositionally, adapt to new domains with minimal supervision, and tightly integrate with formal verification frameworks.

Another important direction involves improving the interactivity and modularity of autoformalization systems. Rather than treating formalization as a purely end-to-end task, future systems may benefit from combining LLMs with symbolic reasoning engines, proof assistants, and logic-based verifiers. This hybrid approach can provide more controllable and interpretable outputs, allowing users to guide or verify parts of the process as needed.

Finally, the long-term vision for autoformalization aligns with broader goals in AI research: to create systems capable of creative mathematical reasoning. Such systems would not only verify existing proofs but also propose conjectures, explore novel ideas, and assist human mathematicians in discovering new research directions \cite{tao2024machine}. Achieving this will require progress at the intersection of natural language understanding, formal logic, mathematical modeling, and human-AI collaboration.

\section{Different with Other Survey}

Existing studies such as \cite{yang2024formal}, \cite{li2024survey}, and \cite{blaauwbroek2024learning} all focus on the development of formal theorem proving in the era of artificial intelligence, outlining various research methods for formal theorem proving. Moreover, \cite{yang2024formal} and \cite{li2024survey} have also summarized and described the model method of autoformalization. Among them, \cite{yang2024formal} divides the autoformalization model methods into rule-based autoformalization and autoformalization based on neural networks and LLMs. The autoformalization model method in this paper draws on this classification approach and further expands the cutting-edge research on this basis.

However, this paper is not limited to the model method itself, but delves deeply into the entire workflow of autoformalization and analyzes it at a more nuanced level. Compared with \cite{yang2024formal} and \cite{li2024survey} which only classify and describe the model methods, this paper goes further on this basis. It not only conducts a detailed interpretation and analysis of the model methods, but also conducts an in-depth study on each part in the autoformalization workflow. Including data preprocessing, model methods, post-processing techniques, result verification, it reveals the internal mechanism of autoformalization from a more subtle perspective.

Furthermore, this paper also combines current hotspots to explore the autoformalization of formal verification for LLMs, and particularly emphasizes the crucial significance of autoformalization for enhancing the credibility and reasoning ability of large language models in the current era of large models. Unlike \cite{blaauwbroek2024learning} which focuses on the development of formal theorem proving, this paper approaches from two unique perspectives: formal verification and mathematical problems, to deeply explore the application and value of autoformalization in LLMs, providing new ideas and methods for improving the performance of LLMs.

\bibliographystyle{named}
\bibliography{ijcai25}

\end{document}